# A Deep Spatial Contextual Long-term Recurrent Convolutional Network for Saliency Detection

Nian Liu and Junwei Han, *Senior Member, IEEE*

*Abstract*—Traditional saliency models usually adopt hand-crafted image features and human-designed mechanisms to calculate local or global contrast. In this paper, we propose a novel computational saliency model, i.e., *deep spatial contextual long-term recurrent convolutional network* (DSCLRCN) to predict where people looks in natural scenes. DSCLRCN first automatically learns saliency related local features on each image location in parallel. Then, in contrast with most other deep network based saliency models which infer saliency in local contexts, DSCLRCN can mimic the cortical lateral inhibition mechanisms in human visual system to incorporate global contexts to assess the saliency of each image location by leveraging the *deep spatial long short-term memory* (DSLSTM) model. Moreover, we also integrate scene context modulation in DSLSTM for saliency inference, leading to a novel *deep spatial contextual LSTM* (DSCLSTM) model. The whole network can be trained end-to-end and works efficiently when testing. Experimental results on two benchmark datasets show that DSCLRCN can achieve state-of-the-art performance on saliency detection. Furthermore, the proposed DSCLSTM model can significantly boost the saliency detection performance by incorporating both global spatial interconnections and scene context modulation, which may uncover novel inspirations for studies on them in computational saliency models.

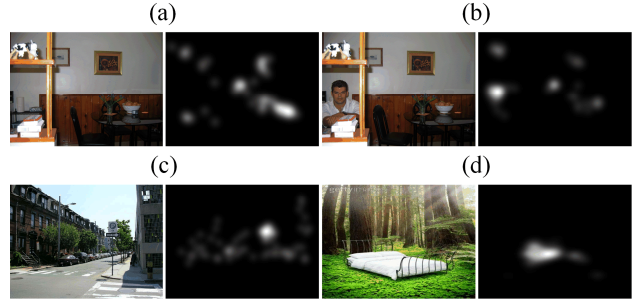

Fig. 1. The influences of global context and scene context to visual saliency. We show 4 pairs of images here. In each pair, the left one is the image stimulus, and the right one is the corresponding ground truth saliency map. Pairs (a) and (b) show the influence of the global context, and pairs (c) and (d) show the influence of the scene context.

## I. INTRODUCTION

WHEN facing visual scenes, human visual system is capable of quickly focusing our eyes on some distinctive visual regions and ignoring plain ones. This neural mechanism is known as visual attention and benefits human beings a lot by helping us quickly and efficiently observing, thinking, and then making decision. There are two forms of visual attention [1]. One is bottom-up saliency-driven attention, which is up to the distinctiveness of visual elements and helps humans to rapidly concentrate on key points of visual scenes. The other one is the top-down task-driven attention, which is driven by endogenous factors, such as one's prior knowledge and how people process their tasks, and helps people to complete the tasks efficiently. In this paper, we focus on the former to predict where people look when free-viewing natural scenes.

In computer science community, researchers have developed lots of computational models to detect visual saliency, most of which follow the biological evidence that salient regions usually stand out from their surroundings and adopt the contrast mechanism to evaluate saliency. Contrast measures the distinctiveness of each image location with respect to a local context or global context, which involves two steps as described below. First, image representations, on which contrast inference is operated, need to be constructed. Traditional methods usually utilize various hand-designed features to represent images, including low-level features, e.g., intensity, color, and orientation [2], middle-level features, e.g., bag of words based shape features and color name features [3], high-level semantic features, e.g., person detection, face detection, car detection [4, 5], and motion features [6]. While effective, these features are manually designed in terms of researcher's domain knowledge on human visual attention, which may be insufficient to simulate the reaction of sophisticated human visual system when facing various natural scenes. Thus, novel and more abundant feature representations are needed as basis for contrast inference.

Second, contrast inference is executed based on the extracted image representations to evaluate saliency. Most traditional methods adopt local contrast, i.e., assessing the difference between each image location and its local surroundings, to predict saliency. For instance, Itti *et al.* [2] computed the local center-surround differences between "center" fine scales and "surround" coarser scales on three feature channels. However, cortical lateral inhibition mechanisms in human visual system

This work was supported in part by the National Science Foundation of China under Grant 61473231 and Grant 61522207. *(Junwei Han is the corresponding author)*

N. Liu and J. Han are with School of Automation, Northwestern Polytechnical University, China (e-mail: junweihan2010@gmail.com).

suggest that neighboring similar features can inhibit each other via specific, anatomically defined interconnections, thus perceived contrast of a centrally viewed test stimulus will be mediated by peripherally viewed flanking stimuli [7]. [8, 9] also presented that visual cortex neurons would compete with each other with the presence of multiple stimuli in the visual scene. All these biological evidences indicate that different spatial locations in visual scenes should be considered holistically in visual attention, instead of only considering local regions in local contrast inference. Thus, global context should also be considered in contrast inference. As shown in Figure 1, the most salient regions in pair (a) are the regions with the stuff on the table. However, when a man appears in the image in pair (b), the most salient region shifts to the man's face. If only local contrast is considered and the inner competition in the visual scene is ignored, the saliency map in pair (b) will probably highlight the region above the table, leading to false positive results. Some methods have already taking global contrast into account by calculating the discrepancy between each image location and the whole image. For instance, Harel *et al.* [10] constructed a fully connected graph over all locations of the image and computed the equilibrium distribution as saliency values. Nevertheless, most previous works resort hand-designed operations or formulations to infer contrast, which may suffer from human's unthorough understanding of the visual attention mechanisms.

To deal with the previously discussed intrinsic problems existed in the saliency detection task, in this paper we propose a novel end-to-end model to detect saliency based on deep neural networks (DNN) as shown in Figure 2. Specifically, we first adopt a deep convolutional neural network (CNN) [11] to extract local image feature representations at each spatial location in parallel. By finetuning deep CNN models [12, 13] pretrained with Imagenet [14] large scale dataset on the saliency detection data, CNNs can automatically learn various saliency-related features hierarchically from raw image data, *e.g.*, color, shape, objects, faces, local contrast, etc. In addition, benefitting from successive convolution and pooling operations, effective local feature maps can be extracted efficiently. Then we utilize *long short-term memory* (LSTM) [15] to model the global context. LSTM is usually used to memorize sufficient context information in time series data via its memory cell. Here we propose to adopt a *deep spatial LSTM* (DSLSTM) model on the obtained convolutional feature map, thus mimicking the human visual system to introduce lateral interconnections among different spatial locations. Supervised by the ground truth eye fixation data, DSLSTM can learn to memorize the long-term spatial interactions, i.e., global context, to evaluate saliency of each image pixel, instead of being restricted in a local context as most traditional works did.

Moreover, scene context can also supply informative hints to visual saliency detection, which has not been deeply studied by most previous works. As one of the few works which studied the role of scene modulation on visual attention, Torralba *et al.* [16] analyzed the gaze distribution over a large annotated image database, i.e., the LableMe dataset [17], and found that eye movements are highly related with scene context. For example, pedestrians are the most salient object in only 10% of the outdoor scene images, being less salient than many other objects. Tables and chairs are among the most salient objects in indoor scenes. Based on these observations, they proposed a Bayesian framework to incorporate scene context in natural search tasks. We also show some intuitive examples in Figure 1. As we can see, being aware of the scene context can help human to quickly focus our eyes on some scene specific important objects (*e.g.*, traffic signs in street views as shown in (c)) or some exceptional objects (*e.g.*, a bed in a forest as shown in (d)). Thus scene context can be seen as an extra top-down high-level semantic factor, as a supplementary to other widely studied top-down object level semantics. Different from [16], in this paper, we try to learn the modulation effect of scene context on attention of free viewing. To be specific, as shown in Figure 2, we first use a state-of-the-art CNN model for scene classification [18] to extract scene features of images, then we embed them as contextual information [19] into the DSLSTM, obtaining a novel *deep spatial contextual LSTM* (DSCLSTM) model, which can simultaneously incorporate global context and scene context information to assess the saliency of each image pixel.

The whole model can be trained end-to-end, including the local image feature extractor CNN, the scene feature extractor CNN, and the DSCLSTM. Then we yield a novel holistic model, i.e., a *deep spatial contextual long-term recurrent convolutional network* (DSCLRCN), to detect visual saliency. When testing, DSCLRCN takes each image as input and directly outputs its saliency map, which is quite straightforward but effective.

In summary, our novelties and contributions are threefold:

1) We propose a novel end-to-end saliency detection model, i.e., the DSCLRCN. Trained with saliency data, it can learn powerful saliency-related local feature representations first, then it learns to simultaneously incorporate global context and scene context to infer saliency.

2) We propose a novel deep spatial contextual LSTM (DSCLSTM) model to effectively learn long-term spatial interactions and scene contextual modulation to infer image saliency. Experiments show that the proposed DSCLSTM can significantly improve saliency detection performance. This may uncover novel insights for future computational saliency models to focus on global and scene contexts.

3) The proposed DSCLRCN model achieves state-of-the-art performance on two benchmark datasets and outperforms other 14 contemporary saliency methods. Furthermore, it also works very efficiently.

The rest of this paper is organized as follows. Section II reviews some works related to our paper. Section III articulates the proposed DSCLRCN model in details. Section IV reports the experimental results on two eye fixation benchmark datasets and the ablation analysis of our model. Finally, we draw conclusion in Section V.

## II. RELATED WORK

Traditional methods usually assess the saliency of each image location with respect to either local contexts or the global context. The former school of methods infer contrast, rarity, or distinctiveness of each image location in local contexts. As one of the earliest pioneer works, Itti *et al.* [2] proposed the "Difference of Gaussians" (DoG) operator to compute the feature difference across Gaussian pyramids of three feature channels, i.e., color, intensity, and orientation, as local contrasts. Then the final saliency map are obtained as an average of the contrast maps. Bruce and Tsotsos [20] computed bottom-up saliency as Shannon's self-information of image features learned by ICA on each local image patch. Gao *et al.* [21] detected salient locations by maximizing the KL divergence between the feature distributions of center and surround regions in an image. Seo and Milanfar [22] built "self-resemblance" maps which measure the center-surround similarity on features based on local regression kernels to assess the likelihood of saliency. Han *et al.* [23] utilized sparse coding model and encoded center locations with dictionaries trained on surrounding locations, then the local contrast can be calculated by combining coding sparseness and residual. Judd *et al.* [4] and Borji [5] extracted low-level and top-down features at each image location and trained classifiers to decide each location to be salient or non-salient. Liang and Hu [3] explored more middle-level features and combined them with object detector features to assess the saliency of each image location.

On the contrary, some other methods resort to the global context, i.e., the saliency of each image pixel is evaluated by considering the whole image. Hou and Zhang [24] transformed the whole image into frequency domain first, then they extracted the spectral residual and transformed it back to spatial domain to obtain saliency map. Zhang *et al.* [25] utilized a Bayesian framework to combine bottom-up saliency with top-down information, then overall saliency emerged as the pointwise mutual information between local image features and the search target's features when performing target searching task. Hou *et al.* [26] proposed the "image signature", which was the sign of the Discrete Cosine Transform (DCT) of an image, as a binary and holistic image descriptor to detect salient image locations. Garcia-Diaz *et al.* [27] proposed to extract local multioriented multiresolution features in Lab color space first, then they performed global whitening normalization on each feature map, and subsequently fused them to obtain the final saliency map.

Recently, benefitting from the great success DNNs achieved on various computer vision tasks [12, 28-30], some researchers also applied DNNs into saliency detection, including salient object detection [31-34] and eye fixation prediction [35-43], and have achieved superior results. Here we mainly focus on eye fixation models. Shen *et al.* [35] and Vig *et al.* [36] used a 3-layer convolutional sparse coding model and hierarchical neuromorphic networks to learn effective image features first, respectively. Then they both adopted a linear SVM to classify each local image location to be salient or non-salient. Han *et al.* [37] first utilized a stacked denoising autoencoder (SDAE) to learn feature representations on sampled image patches. Then another SDAE was used to learn center-surround contrast for saliency inference. Kümmerer *et al.* [38] learned a softmax classifier on linearly combined multi-level features of AlexNet [28] on each image location to predict eye fixations. Liu *et al.* [39] used a multi-resolution CNN to combine multi-scale contexts and do saliency classification on each image location. [40-45] all utilized fully convolutional networks (FCNs) [30] based on pretrained deep networks (i.e., VGGnet [12]) to infer saliency of each image location in parallel. However, all these previous works assessed saliency in local contexts due to their local features and pixel-wise classifiers [35-39] or limited receptive fields in FCN based models [40-45]. Although some works [41, 42] tried to capture global context using convolutional layers with very large receptive field, this idea only held on a few image locations around the image center while failed on other positions due to the intrinsic property of convolutional layers.

We propose to adopt DSLSTM to construct interconnections among different image locations to incorporate long term global context. *Recurrent neural networks* (RNNs), including their improved variant LSTM, have shown their excellent capability to memorize long term contexts in time series data, e.g., speech recognition [46] and natural language processing [47]. Lately, RNNs have also been applied into computer vision tasks. Donahue *et al.* [48] proposed *Long-term Recurrent Convolutional Networks* (LRCN) which stacked LSTM on temporal dimension on CNN encoder features to deal with video recognition and image description tasks. Visin *et al.* [49] proposed the ReNet model in which four recurrent neural networks swept horizontally and vertically in both directions across the image to learn context features for image classification. Bell *et al.* [50] and Yan *et al.* [51] applied the ReNet model on top of CNN features to integrate context information for object detection and semantic segmentation, respectively. Based on the ReNet model, in this paper we adopt DSLSTM with concatenation with deep CNNs to incorporate global context for saliency detection.

Moreover, Ghosh *et al.* [19] proposed the *Contextual LSTM* (CLSTM) model to incorporate topics as contextual information into LSTM for NLP tasks. While we propose to embed scene features as contextual information into the DSLSTM model, obtaining a novel DSCLSTM model, to simultaneously incorporate global context and scene modulation for saliency detection.

## III. DSCLRCN FOR SALIENCY DETECTION

In this section, we illuminate the proposed DSCLRCN in details for saliency detection. Specifically, as shown in Figure 2, we first adopt a pretrained CNN model to extract local convolutional image features. At the same time, the preatrained Places-CNN [18] model is also used to extract a scene feature vector. Then local image features and the scene vector are both normalized and fed into the DSCLSTM, which propagates the global and scene contextual information to each image location. Finally, the saliency map can be obtained by a simple convolutional layer, and the NSS loss between the upsampled saliency map and the human eye fixation locations is used as supervision to train the whole network. Below we describe each network component in details.

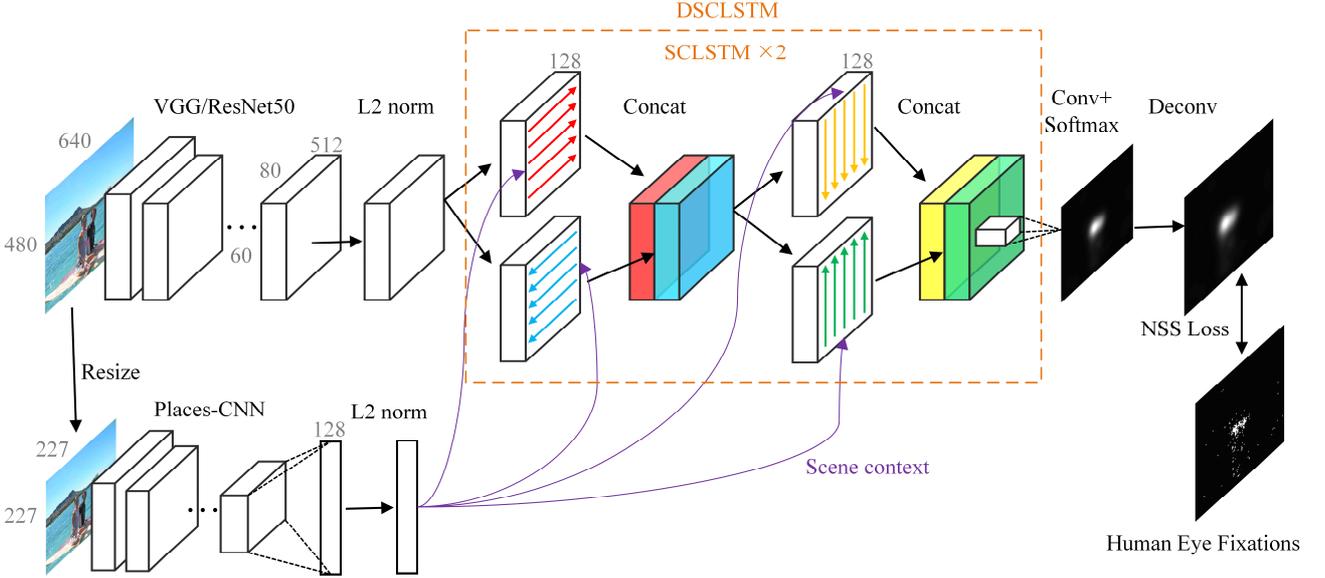

Fig. 2. The network architecture of the proposed DSCLRCN. First, local feature map and scene feature are extracted using pretrained CNNs. Then, a DSCLSTM model is adopted to simultaneously incorporate global context and scene context. Finally, saliency map is generated and upsampled.

*A. Local Image Feature Extraction using CNNs*

We adopt deep CNNs pretrained on Imagenet [14] dataset to extract local image features using fully convolutional architecture [30]. The network is based on VGG 16-layer network [12] or the ResNet-50 network [13]. To preserve a relatively large size of the extracted feature map, we utilize the dialated convolution [52] strategy, which supports exponential expansion of the receptive field without loss of resolution or coverage.

VGG 16-layer network consists of 13 convolutional layers with $3\times3$ convolutional kernels and 3 fully connected layers. To maintain the spatial information, we only utilize the convolutional layers, which are composed of five convolutional blocks and each of them is followed by a max-pooling layer with downsampling stride 2. We keep the layers before the forth convolutional block (conv4) as the same with the original VGG network. After that we discard pool4 and pool5 layers and adopt dilation sizes [52] of 2 in conv5 layers to preserve the resolution and receptive field sizes. To enlarge the receptive filed size of the neurons in the final feature map, we add another two convolutional blocks each of which consists of two convolutional layers with 512 $3\times3$ convolutional kernels, dilation sizes of 4, and ReLU [53] activation function. Since the activation values after ReLU activation in the last layer are usually very large, which will make the hidden neurons in subsequent LSTM layers easily saturate and hard to train, we use the $L_2$-norm layer [54] to normalize the whole feature map to have standard $L_2$-norm first and then learn to re-scale it to an appropriate magnitude for the subsequent LSTM layers. The overall network structure is shown in Table I.

The ResNet-50 network consists of 49 convolutional layers and 1 fully connected layer. Once again, we only use the convolutional layers to extract local image features, which consist of 5 blocks of convolutional layers. The first block is just one convolutional layer with stride 2, followed by a max-pooling layer with stride 2, either. As for other 4 blocks, each of them is composed of several residual learning blocks [13] and all the last 3 blocks have strides 2. Similarly, we keep the layers before the conv3 block and revise the conv4 and conv5 blocks to have strides 1 and dilation sizes of 2 and 4, respectively. Because the feature map in the last layer has relatively large channel numbers (2048), we utilize a convolutional layer with ReLU activation function to reduce the dimension to a relatively small one (we set it to 512 in this work) for easy learning of subsequent LSTM layers. Unlike the ResNet layers, we do not use batch normalization [55] in this layer. Finally, we again use the $L_2$-norm layer to normalize the feature map and rescale it. The overall network structure is shown in Table II.

We also consider multilayer features for the ResNet50 model to incorporate multiscale contexts. Specifically, we use both of the conv4 feature and conv5 feature. First, we use two convolutional layers with 512 $1\times1$ convolutional kernels and ReLU activation function on top of conv4 and conv5 feature maps to reduce their feature channels. Subsequently, two $L_2$-norm layers with same scale parameters are used to make the two feature maps compatible. At last, they are concatenated and another channel-reduction convolutional layer is adopted to obtain the final feature map with 512 channels.

Both of the two networks have strides of 8. Thus when we feed an image with size $P\times Q$ into the feature extractor CNNs, we can yield a convolutional feature map of size $\frac{P}{8}\times\frac{Q}{8}\times 512$, which will be inferred as the *local feature map* below.

As for the scene feature extractor CNN, we first simply resize each image to size $227\times227$ and adopt the convolutional layers of Places-CNN [18] model to extract convolutional features. Then we add one fully connected layer with 128 neurons and ReLU activation function on top of the pool5 feature. Finally we use another $L_2$-norm layer to normalize the feature of the fully connected layer and rescale it. Thus we obtain the *scene feature*, which will be inputted into the subsequent DSCLSTM model with the local feature map.

TABLE I
ARCHITECTURE OF THE VGG BASED LOCAL FEATURE EXTRACTOR. CONVOLUTIONAL SETTINGS ARE GIVEN BY [CHANNEL,KERNEL_STRIDE_DILATION]× LAYERS. THE POOLING SETTINGS ARE GIVEN BY [KERNEL_STRIDE].

| Name | conv1 | pool1 | conv2 | pool2 | conv3 | pool3 | conv4 | conv5 | conv6 | conv7 | norm |
|---|---|---|---|---|---|---|---|---|---|---|---|
| Setting | [64,3_1_1]×2 | [2_2] | [128,3_1_1]×2 | [2_2] | [256,3_1_1]×3 | [2_2] | [512,3_1_1]×3 | [512,3_1_2]×3 | [512,3_1_4]×2 | [512,3_1_4]×2 | $L_2$-norm scale:400 |

TABLE II
ARCHITECTURE OF THE RESNET50 BASED LOCAL FEATURE EXTRACTOR. CONVOLUTIONAL SETTINGS ARE GIVEN BY [CHANNEL,KERNEL_STRIDE_DILATION]× LAYERS. THE POOLING SETTINGS ARE GIVEN BY [KERNEL_STRIDE]. BATCH NORMALIZATION [54] IS USED IN RESNET LAYERS (CONV1 TO CONV5).

| Name | conv1 | pool1 | conv2 | conv3 | conv4 | conv5 | conv6 | norm |
|---|---|---|---|---|---|---|---|---|
| Setting | [64,7_2_1]×1 | [3_2] | $\begin{bmatrix}64,1\_1\_1\\64,3\_1\_1\\256,1\_1\_1\end{bmatrix}\times 3$ | $\begin{bmatrix}128,1\_2\_1\\128,3\_1\_1\\512,1\_1\_1\end{bmatrix}$ + $\begin{bmatrix}128,1\_1\_1\\128,3\_1\_1\\512,1\_1\_1\end{bmatrix}\times 3$ | $\begin{bmatrix}256,1\_1\_1\\256,3\_1\_2\\1024,1\_1\_1\end{bmatrix}\times 6$ | $\begin{bmatrix}512,1\_1\_1\\512,3\_1\_2\\2048,1\_1\_1\end{bmatrix}\times 3$ | [512,1_1_1]×1 | $L_2$-norm scale:400 |

### B. DSCLSTM for Context Incorporation

In this section we introduce the proposed DSCLSTM model which can simultaneously incorporate global context information and scene context modulation. We first briefly review the LSTM model, then we elaborate the DSLSTM model and how to embed scene context into it.

*1) Reviewing LSTM*

LSTM is a variant of RNNs and was proposed by Hochreiter and Schmidhuber [15] to solve the vanishing gradient problem by introducing a *memory cell* to keep states over long-term time series data. At each time step $t$, a LSTM unit has a memory cell $c_t \in R^N$ and the hidden state $h_t \in R^N$, where $N$ is the number of hidden units. Given the input $x_t \in R^M$ with input dimension $M$, the previous memory cell $c_{t-1}$, and the previous hidden state $h_{t-1}$, LSTM unit updates its memory cell and hidden state via four gates, namely, input gate $i_t \in R^N$, forget gate $f_t \in R^N$, output gate $o_t \in R^N$, and input modulation gate $g_t \in R^N$. The transition equations are given by:

$$i_t = \sigma(\mathbf{W}_{xi}x_t + \mathbf{W}_{hi}h_{t-1} + b_i), \quad (1)$$
$$f_t = \sigma(\mathbf{W}_{xf}x_t + \mathbf{W}_{hf}h_{t-1} + b_f), \quad (2)$$
$$o_t = \sigma(\mathbf{W}_{xo}x_t + \mathbf{W}_{ho}h_{t-1} + b_o), \quad (3)$$
$$g_t = \phi(\mathbf{W}_{xc}x_t + \mathbf{W}_{hc}h_{t-1} + b_c), \quad (4)$$
$$c_t = f_t \odot c_{t-1} + i_t \odot g_t, \quad (5)$$
$$h_t = o_t \odot \phi(c_t), \quad (6)$$

where $\sigma$ and $\phi$ are element-wise sigmoid and hyperbolic tangent function, respectively. $\odot$ represents element-wise multiplication. $\mathbf{W}_*$ and $b_*$ are learnable weights and biases, which can be trained by backpropagation through time (BPTT) algorithm [56]. Finally, we can represent the whole process as:

$$(h_t, c_t) = LSTM(x_t, h_{t-1}, c_{t-1}). \quad (7)$$

As we can see, the forget gate $f_t$ determines the amount of the previous memory cell $c_{t-1}$ to be kept by $c_t$. $g_t$ can be seen as a preactivation of $c_t$ which is contributed by current input signal, while the input gate $i_t$ controls how much information of $g_t$ are permitted to input into $c_t$ and update it. While being modulated by both $f_t$ and $i_t$, the memory cell $c_t$ learns to selectively forget or memorize previous memory and current signal and propagate them to the next time step, thus being capable of incorporating long-term and complex contextual dependencies. Finally, the output gate $o_t$ controls the information flow from current memory $c_t$ to the hidden state $h_t$.

One of the most important variants of LSTM is the *Bidirectional LSTM* (BLSTM) [57]. BLSTM consists of two parallel LSTMs to separately scan the input data sequentially and reversely. Then the hidden states of the two LSTMs are concatenated or added as the one of the BLSTM, which captures both past and future information.

*2) DSLSTM for Global Context Incorporation*

Now we introduce the spatial LSTM (SLSTM) based on the ReNet model [49] for incorporating global context into saliency detection. SLSTM takes the local feature map extracted previously as the input, then it runs four LSTMs operating in different directions over the feature map, i.e., two BLSTMs scanning horizontally and vertically, to blend the context information. Specifically, as shown in the orange box in Figure 2, it first treats each pixel in each row of the feature map as a time step and runs two LSTMs in parallel to scan it left-to-right and right-to-left, thus obtaining two feature maps which consist of the hidden states of the LSTMs at each spatial location. Let us represent the input local feature map as $\mathbf{X}$, thus at each location $(p,q)$ we have $x_{p,q} \in R^M$, where $M = 512$, $p = 1, \cdots, \frac{P}{8}$, and $q = 1, \cdots, \frac{Q}{8}$ as mentioned in Section A. Then the updating of the cell memory $c_{p,q}$ and the hidden state $h_{p,q}$ at location $(p,q)$ can be represented by:

$$(h_{p,q}^{\rightarrow}, c_{p,q}^{\rightarrow}) = LSTM^{\rightarrow}(x_{p,q}, h_{p,q-1}^{\rightarrow}, c_{p,q-1}^{\rightarrow}),$$
$$(h_{p,q}^{\leftarrow}, c_{p,q}^{\leftarrow}) = LSTM^{\leftarrow}(x_{p,q}, h_{p,q+1}^{\leftarrow}, c_{p,q+1}^{\leftarrow}),$$
(8)

where the signs $\rightarrow$ and $\leftarrow$ represents the left-to-right and right-to-left scanning, respectively.

Next the SLSTM concatenates the two feature maps $\mathbf{H}^{\rightarrow}$ and $\mathbf{H}^{\leftarrow}$ along the channel dimension, obtaining the fused feature map $\mathbf{H}^{\leftrightarrow}$, which incorporates the context information from both left and right at each location.

Then, the SLSTM uses another two LSTMs to scan $\mathbf{H}^{\leftrightarrow}$ from top to bottom and bottom to top, which can be represented by:

$$(h_{p,q}^{\downarrow}, c_{p,q}^{\downarrow}) = LSTM^{\downarrow}(h_{p,q}^{\leftrightarrow}, h_{p-1,q}^{\downarrow}, c_{p-1,q}^{\downarrow}),$$
$$(h_{p,q}^{\uparrow}, c_{p,q}^{\uparrow}) = LSTM^{\uparrow}(h_{p,q}^{\leftrightarrow}, h_{p+1,q}^{\uparrow}, c_{p+1,q}^{\uparrow}).$$
(9)

The signs $\downarrow$ and $\uparrow$ represent the top-to-bottom and bottom-to-top scanning, respectively.

At last, SLSTM concatenates $\mathbf{H}^{\downarrow}$ and $\mathbf{H}^{\uparrow}$, obtaining $\mathbf{H}^{\updownarrow}$. By progressively scanning the local feature map horizontally and vertically in four directions, the information at each location can be propagated to any other locations and each location in $\mathbf{H}^{\updownarrow}$ contains contextual interactions from all other locations, thus long-term global contextual dependencies are incorporated in a very efficient way.

We stack two SLSTMs successively, leading to DSLSTM. The increased depth is supposed to increase the capability to learn longer-range and more complex contextual dependencies between different locations. Experiments also show that deep SLSTMs are more effective to blend contextual information in the whole image in section IV.

*3) Scene Context Modulation: DSCLSTM*

Inspired by the CLSTM model in [19] and the fact that the scene modulation effects in visual attention, we propose to embed the scene feature into the DSLSTM to integrate scene contexts in our saliency model. Specifically, [19] inputted a topic vector into the traditional LSTM unit at each time step as a static input, i.e., added linear projections of the topic vector to the formulations of the four gates (Equations (1) to (4)). While we find that adding a static input into each time step can easily lead to poor local minimum when training the network, thus we just add the scene feature $s$ in the first time step, where we have:

$$i_1 = \sigma(\mathbf{W}_{xi} x_1 + \mathbf{W}_{hi} h_0 + \mathbf{W}_{si} s + b_i), \quad (10)$$
$$f_1 = \sigma(\mathbf{W}_{xf} x_1 + \mathbf{W}_{hf} h_0 + \mathbf{W}_{sf} s + b_f), \quad (11)$$
$$o_1 = \sigma(\mathbf{W}_{xo} x_1 + \mathbf{W}_{ho} h_0 + \mathbf{W}_{so} s + b_o), \quad (12)$$
$$g_1 = \sigma(\mathbf{W}_{xc} x_1 + \mathbf{W}_{hc} h_0 + \mathbf{W}_{sc} s + b_c). \quad (13)$$

We adjust the four types of LSTM in the DSLSTM, i.e., $LSTM^{\rightarrow}$, $LSTM^{\leftarrow}$, $LSTM^{\downarrow}$, and $LSTM^{\uparrow}$ accordingly to incorporate the scene contexts from their first time steps. However, benefitting from the memory cells, the scene contextual information can still be propagated to other time steps, i.e., the whole feature map, leading to the novel DSCLSTM model.

The output feature map of DSCLSTM has $2N$ channels and the same spatial size as the local feature map. However, the feature at each location has simultaneously integrated the global context and scene modulation. Now, the features are ready for saliency assessment.

*C. Saliency Assessment*

We simply adopt a convolutional layer with 1 1×1 kernel and the Softmax activation function to generate the saliency map. The Softmax activation function is used to normalize the whole map, thus introducing lateral competition for saliency assessment. Because the saliency map is generated by a stride of 8, subsequently we use a deconvolutional layer with bilinear interpolation kernels [30] to upsample the saliency map with a stride of 8. Thus, we can obtain a saliency map with the same size as the input image.

When training, we use the negative Normalized Scanpath Saliency (NSS) [58] of the saliency map with reference to the corresponding ground truth human eye fixations as the objective function to train the network. NSS is chosen for the recommendation of [59] and will be elaborated later. The whole DSCLRCN model can be trained end-to-end using back propagation algorithm [60].

When testing, we just feed each testing image into DSCLRCN and can directly yield the saliency map, being straightforward but effective.

IV. EXPERIMENTS

In this section, we report experimental results to evaluate the effectiveness of DSCLRCN in the saliency detection task. We first introduce the eye fixation benchmark datasets and evaluation metrics we used in this work. Then we do model ablation analysis to evaluate the contribution of each model component. Finally we compare DSCLRCN with other state-of-the-art methods both quantitatively and qualitatively to show the effectiveness of our proposed model.

*A. Datasets*

We evaluated DSCLRCN on two benchmark datasets. The first one is **SALICON** [61], i.e., Saliency in Context dataset, which contains 10,000 training images, 5,000 validation images, and 5,000 testing images. The images are all of size

480×640 and selected from the MS COCO [62] dataset with rich contextual information. The ground truth of eye fixations are collected by a proposed mouse-contingent multi-resolutional paradigm and have been shown to be highly similar with eye tracking data. Benefitting from the novel paradigm, this dataset, as the largest eye fixation dataset so far, can be created much efficiently and easily. The ground truth of eye fixations of the testing set are held-out and researchers are supposed to submit their results to the SALICON challenge website[1] or the LSUN Saliency Challenge website[2] to evaluate their methods.

The second dataset for evaluation is **MIT300** [63]. It contains 300 images with natural outdoor or indoor scenes, and has become one of the widely used benchmark datasets in recent years. The ground truth of eye fixation data are held-out and researchers can submit their models to the MIT Saliency Benchmark website [3] to evaluate their models. As the organizers suggested, the **MIT1003** [4] dataset can be used as the training and validation sets for MIT300 since they are collected with similar eye-tracking setup. This dataset contains 779 landscape images and 228 portrait images collected from Flickr and LabelMe, and the eye fixation data are collected while being viewed by 15 human subjects.

### B. Evaluation Metrics

There exist various evaluation metrics for eye fixation prediction, including Earth Mover's Distance (**EMD**), Normalized Scanpath Saliency (**NSS**), Pearson's Correlation Coefficient (**CC**), Similarity (**SIM**), Area Under Curve (**AUC**), shuffled-AUC (**sAUC**), Kullback-Leibler divergence (**KL**), Information Gain (**IG**), etc. Bylinskii *et al.* [59] showed that among these metrics, KL, IG, and SIM are most sensitive to false negatives, AUC metrics ignore low-valued false positives, EMD's penalty depends on spatial distance, while NSS and CC are equally affected by false positives and negatives. Thus based on their recommendation, we report CC and NSS here. Moreover, AUC and shuffled-AUC are also reported for comparison with existed models for historical reasons.

The Area Under the ROC Curve (**AUC**) metric is widely used to evaluate saliency models. For an image with its binary ground truth eye fixation map $\mathbf{G}^F$, AUC evaluates the classification performance of the computed saliency map $\mathbf{S}$, where fixation points and non-fixation points in $\mathbf{G}^F$ are considered as the positive set and negative set, respectively. Specifically, $\mathbf{S}$ is normalized to [0, 1] first. Then it is binarily classified into salient regions and non-salient regions by a threshold. By varying the threshold from 0 to 1, ROC curves can be obtained by plotting true positive rate vs. false positive rate. Finally, the area under the ROC curve is calculated as the AUC score. To alleviate the influence of center-bias, [25, 64] introduced **sAUC**, which adopts the fixation points of other images in the dataset as the negative set. Although widely used, AUC metrics are ambivalent to monotonic transformations and ignore low-valued false positives, which may be unfavorable behavior for eye fixation prediction [59].

**NSS** is introduced in [58], which computes the average of the normalized saliency values at eye fixation locations. As analyzed in [59], it is sensitive to false positives, relative differences in saliency across the image, and monotonic transformations. Given a saliency map $\mathbf{S}$ and the corresponding eye fixation map $\mathbf{G}^F$, NSS is calculated as:

$$NSS(\mathbf{S},\mathbf{G}^F) = \frac{1}{\sum \mathbf{G}^F} \sum \overline{\mathbf{S}} \odot \mathbf{G}^F, \qquad (14)$$
$$\overline{\mathbf{S}} = \frac{\mathbf{S}-\mu(\mathbf{S})}{\sigma(\mathbf{S})}.$$

**CC** considers the saliency map $\mathbf{S}$ and the corresponding ground truth saliency map $\mathbf{G}^S$, which is obtained by Gaussian blurring $\mathbf{G}^F$, as random variables and computes their Pearson's Correlation Coefficient:

$$CC(\mathbf{S},\mathbf{G}^S) = \frac{cov(\mathbf{S},\mathbf{G}^S)}{\sigma(\mathbf{S})\times\sigma(\mathbf{G}^S)}. \qquad (15)$$

### C. Implementation Details
#### 1) Data Processing

For the SALICON benchmark dataset, we directly used the SALICON training and validation datasets to train and validate DSCLRCN, respectively. For the MIT300 benchmark dataset, we used 903 images from MIT1003 dataset to finetune the model trained on SALICON. Another 100 images were used for validation. Since MIT1003 contains relatively less images, we augmented the training set twice by using horizontal flipping. For simplicity, we directly resized all images to size 480×640 and 227×227 for the local feature extractor CNN and the scene feature extractor CNN, respectively.

When testing, we also resized testing images to size 480×640 and 227×227 and feedforward them through the network to obtain saliency maps. Then we resized them to the same sizes with the input images. Finally, we used small Gaussian filters to blur the saliency maps. Via validation experiments, for each image we set the standard deviation of Gaussian filters to be $\sigma = 0.035\min(P,Q)$, and set the size of the Gaussian filters to be $4\sigma$.

#### 2) Network Settings

We used stochastic gradient descent (SGD) with momentum to train the whole network. The batchsize was set to 20. For the SALICON dataset, the learning rates of the pretrained layers and other layers were set to 0.001 and 0.01, respectively. We also scaled down the learning rates by a factor of 2.5 every 500 iteration steps. The overall iteration step was set to 5,000, and we validated the trained models every 500 steps to select a best model for testing. While using MIT1003 data to finetune the model trained on SALICON, we set the learning rates for all layers to 0.001 and scaled them down every 100 iteration steps. The overall iteration step and the validation step were set to 1000 and 100, respectively. We also used momentum of 0.9 and a weight decay of 0.0005.

---

[1] https://competitions.codalab.org/competitions/3791
[2] http://lsun.cs.princeton.edu/2016/
[3] http://saliency.mit.edu/

TABLE III
MODEL ABLATION ANALYSIS ON THE SALICON VALIDATION DATASET.
THE BEST SCORE OF EACH METRIC IS SHOWN IN BOLD FACE.

| Settings | sAUC | AUC | NSS | CC |
|---|---|---|---|---|
| influence of the RF sizes | | | | |
| FCN5 (RF: 196) | 0.789 | 0.869 | 2.960 | 0.756 |
| FCN6 (RF: 340) | **0.792** | 0.877 | 3.075 | 0.793 |
| FCN7 (RF: 468) | 0.790 | 0.878 | 3.087 | 0.795 |
| effectiveness of global context incorporation | | | | |
| FCN7_SLSTM | 0.79 | 0.882 | 3.143 | 0.811 |
| FCN7_DSLSTM | 0.786 | 0.883 | 3.160 | 0.816 |
| effectiveness of scene modulation | | | | |
| FCN7_DSCLSTM | 0.786 | 0.884 | 3.171 | 0.822 |
| results with ResNet50 model | | | | |
| ResNet50_DSCLSTM | 0.791 | 0.886 | 3.216 | 0.831 |
| ResNet50_ML_DSCLSTM | 0.788 | **0.887** | **3.221** | **0.835** |

TABLE IV
COMPARISON RESULTS ON THE SALICON TEST DATASET. THE BEST
SCORE OF EACH METRIC IS SHOWN IN BOLD FACE.

| Models | sAUC | AUC | NSS | CC |
|---|---|---|---|---|
| Shallow Convnet [43] | 0.658 | 0.821 | 1.663 | 0.562 |
| Deep Convnet [43] | 0.724 | 0.858 | 1.859 | 0.622 |
| DeepGaze II [38] | **0.787** | 0.867 | 1.271 | 0.479 |
| ML-Net [45] | 0.768 | 0.866 | 2.789 | 0.743 |
| SU [42] | 0.760 | 0.880 | 2.610 | 0.780 |
| DSCLRCN | 0.776 | **0.884** | **3.157** | **0.831** |

TABLE V
COMPARISON RESULTS ON THE MIT300 DATASET. WE USE THE
AUC-JUDD IMPLEMENTATION AS THE AUC METRIC. THE BEST SCORE OF
EACH METRIC IS SHOWN IN BOLD FACE.

| Models | sAUC | AUC-Judd | NSS | CC |
|---|---|---|---|---|
| GBVS [10] | 0.63 | 0.81 | 1.24 | 0.48 |
| Judd [4] | 0.60 | 0.81 | 1.18 | 0.47 |
| AWS [27] | 0.68 | 0.74 | 1.01 | 0.37 |
| BMS [66] | 0.65 | 0.83 | 1.41 | 0.55 |
| eDN [36] | 0.62 | 0.82 | 1.14 | 0.45 |
| Mr-CNN [39] | 0.69 | 0.79 | 1.37 | 0.48 |
| SALICON [40] | 0.74 | **0.87** | 2.12 | 0.74 |
| DeepFix [41] | 0.71 | **0.87** | 2.26 | 0.78 |
| Shallow Convnet [43] | 0.63 | 0.80 | 1.47 | 0.56 |
| Deep Convnet [43] | 0.69 | 0.83 | 1.51 | 0.58 |
| DeepGaze II [38] | **0.76** | **0.87** | 1.29 | 0.51 |
| PDP [44] | 0.73 | 0.85 | 2.05 | 0.70 |
| ML-Net [45] | 0.70 | 0.85 | 2.05 | 0.67 |
| DSCLRCN | 0.72 | **0.87** | **2.35** | **0.80** |

When using the ResNet50 model, to facilitate training, we fixed the scale and bias parameters of the BN layers, i.e., we directly used the global statistics of the original ResNet50 parameters. The initial scales of the $L_2$-norm layers for the local feature map and the scene feature were set to 400 and 9, respectively. The biases of the LSTM forget gates were initialized to 1 to help to keep long-term memories. The parameters in each BLSTM, i.e., $LSTM^{\rightarrow}$ and $LSTM^{\leftarrow}$, $LSTM^{\downarrow}$ and $LSTM^{\uparrow}$, were shared by considering the symmetry of images.

We implemented DSCLRCN using caffe [65] library. The testing code was implemented using Matlab. A GTX Titan X GPU was used both in training and testing for acceleration. The running time for testing an image is 0.27s. The code will be publicly available on the author's homepage[4].

*D. Model Ablation Analysis*

In this section we do ablation analysis on the SALICON validation dataset to evaluate the contribution of each model component. The results are shown in Table III.

*1) Influence of the Receptive Field size*

The size of Receptive field (RF) determines how large the area is which is involved in the activation of a neuron in a CNN layer. We show the evaluation results of using conv5, conv6, and conv7 features of the VGG based feature extractor (shown in Table I) to directly detect saliency in the FCN architecture without DSCLSTM, which are represented by **FCN5**, **FCN6**, and **FCN7** in Table III, respectively. The RF sizes of corresponding feature maps are also given. We can see that, while the RF sizes are enlarged, the saliency detection performance can be boosted by incorporating more context information. Especially, the performance gains significantly from FCN5 to FCN6, which may be attributed to FCN5's relatively too small RF size with respect to the image size

---

[4] https://sites.google.com/site/liunian228/

($196\times196$ vs. $480\times640$). This indicates that saliency detection heavily relies on large contexts.

*2) Effectiveness of Global Context Incorporation*

We directly added SLSTM layers on top of the FCN7 network to evaluate the effectiveness of incorporating the global context. As shown in Table III, adding a SLSTM layer can improve the performance much, while 2 SLSTM layers (DSLSTM) bring more improvement, demonstrating that integrating global context can significantly benefit saliency detection performance. While we didn't observed more meaningful improvements by continuing to deepen SLSTM layers.

*3) Effectiveness of Scene Modulation*

We added scene modulation to the FCN7_DSLSTM model to evaluate its effectiveness. By comparing the performance of FCN7_DSLSTM and FCN7_DSCLSTM in Table III, we can see that incorporating scene context can also improve the saliency detection performance much. This indicates that scene context can also supply much informative information to saliency detection, which deserves more attention and research in the future.

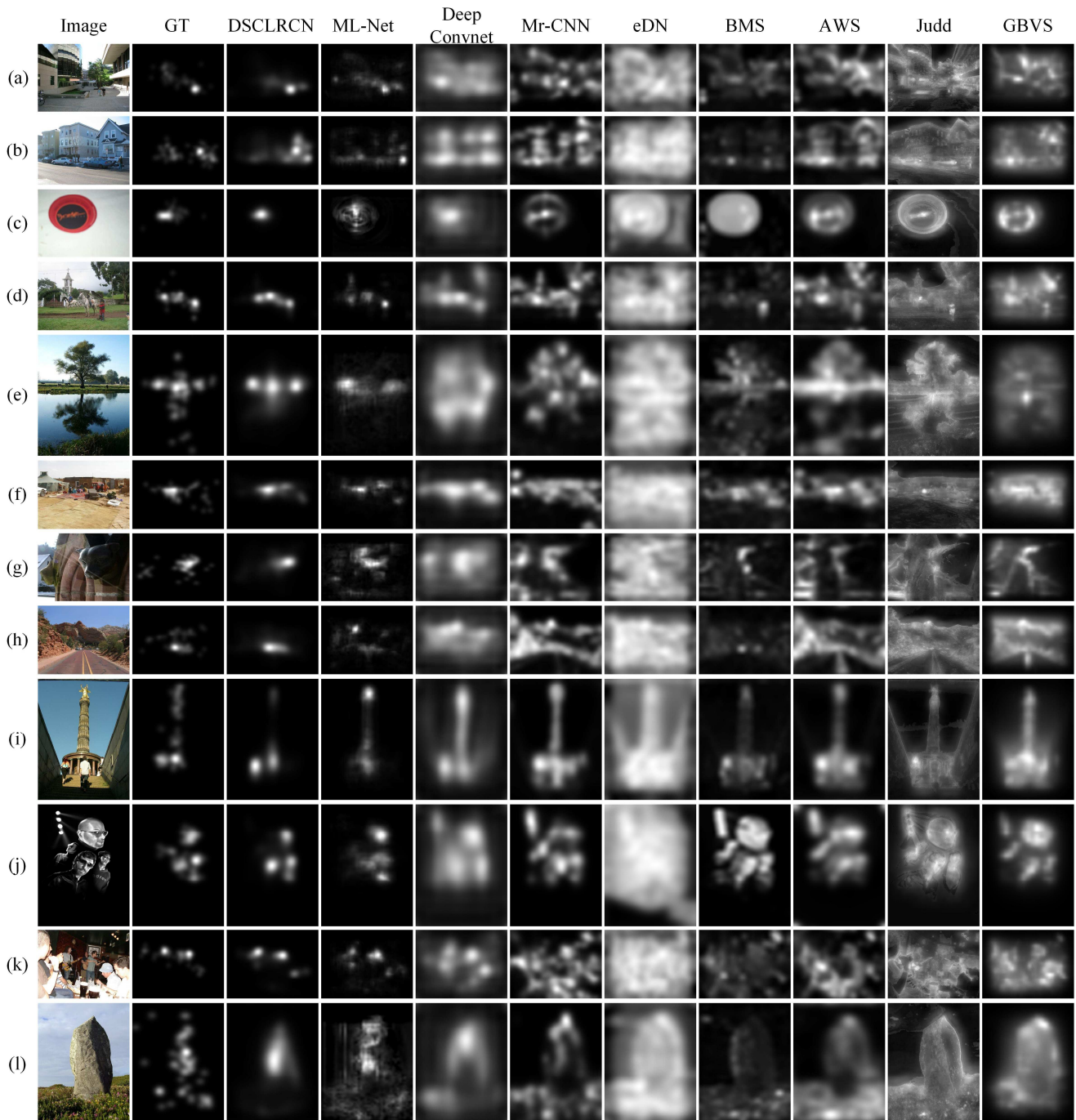

Fig. 3. Qualitative comparison results on the MIT1003 validation set. The second column shows ground truth saliency maps (GT).

*4) Boosting the Performance using Deeper and Multiscale Features*

We show the performance using the ResNet50 based feature extractor network. From Table III, we can see that, using the more powerful ResNet50 feature can further improve saliency detection performance, which is consistent with other observations on other computer vision tasks. We also show the results of using multilayer features of the ResNet50 network (shown as ResNet50_ML_DSCLSTM in Table III). We can see that, with similar conclusions in [31, 40, 41, 45], integrating multiscale features can also further improve saliency detection performance.

We also tried to use multilayer features in the FCN7_DSCLSTM model but obtained worse results. This may be attributed to that in the FCN7 based model, the layers FCN6 and FCN7 were trained from the scratch, thus adding multilayer connections may degrade the network training.

*E. Comparison with state-of-the-arts*

In this section, we compared the proposed DSCLRCN model on the SALICON test dataset and the MIT300 dataset with other 14 state-of-the-art saliency detection models, including GBVS [10], Judd [4], AWS [27], BMS [66], eDN [36], Mr-CNN [39], SALICON [40], DeepFix [41], Shallow Convnet [43], Deep Convnet [43], DeepGaze II [38], PDP [44], SU (Saliency Unified) [42], and ML-Net [45]. It is worth noting that all the last 10 models are proposed recently and are based on DNNs. All these methods (including ours) have submitted their results to the SALICON challenge website or the MIT Saliency Benchmark challenge website, where all the saliency scores are obtained. We used the ResNet50_ML_DSCLSTM setting as our final model due to its best performance.

We show the comparison results on the SALICON test dataset in Table IV. This dataset is recently proposed, thus only some recent models provided results on this dataset, but all of them are DNN based models. We can see that DSCLRCN outperforms all of them in terms of three metrics, i.e., AUC, NSS, and CC. Especially on NSS and CC, DSCLRCN obtains significant superiority, demonstrating its effectiveness. We notice that DeepGaze II [38] achieves better sAUC scores than our method. However, sAUC primarily rewards true positives while being hard to be degraded by false positives [41, 59, 67]. This can usually lead to good score to very blurred/hazy saliency maps [68], which is the case of DeepGaze II.

The comparison results on the MIT300 dataset is shown in Table V. We can see that in generally DSCLRCN outperforms all other previous models, especially including 9 DNN based models. Specifically, DSCLRCN achieves state-of-the-art performance in terms of AUC, and outperforms all other models on NSS and CC, indicating that DSCLRCN generates more accurate highlights and less false positives highlights. We can also see that DSCLRCN outperforms DeepFix [41], which used large convolutional kernels to integrate contextual information. This demonstrates the effectiveness of the proposed DSCLSTM model in incorporating global and scene contexts.

We also show qualitative comparison results on the MIT1003 validation set in Figure 3. We can see that the saliency maps generated by DSCLRCN match the ground truth saliency maps best among all the compared models. Specifically, DSCLRCN generates more accurate detections and much less false positives compared with other models, including 4 DNN based models, i.e., ML-Net [45], Deep Convnet [43], Mr-CNN [39], and eDN [36]. It is worth noting that DSCLRCN can deal with various very challenging scenarios, including cluttered scenes (rows (a), (b), (d), (f), and (k)), and scenes with no obvious salient regions (rows (e), (h), and (l)). These two scenarios are usually very challenging for other models. The former often leads to severe false positive highlights and the latter usually leads to inaccurate highlights. While benefitting from the integrated global and scene contexts, DSCLRCN can accurately detect the most salient regions in an image while ignoring other local distractions, thus obtaining much better results.

V. CONCLUSION

In this paper, we proposed a novel end-to-end saliency model, i.e., DSCLRCN, to predict human eye fixation points in natural scenes. Specifically, DSCLRCN first learned various saliency related local features via finetuning pretrained CNN models. Next, it leveraged the DSLSTM model to incorporate global contexts via mimicing the cortical lateral inhibition mechanisms in human visual system. Furthermore, we also proposed to integrate scene modulation in saliency detection, leading to the novel DSCLSTM model. Experimental results showed that DSCLRCN outperforms all previous saliency models on two eye fixation benchmark datasets, including 10 recently proposed DNN based models. Ablation analysis also showed that the proposed DSCLSTM model can significantly improve saliency detection performance, which may supply new inspirations for future computational saliency models to focus on global and scene contextual information analysis.

REFERENCES

[1] C. E. Connor, H. E. Egeth, and S. Yantis, "Visual attention: bottom-up versus top-down," *Current Biology,* vol. 14, no. 19, pp. R850-R852, 2004.

[2] L. Itti, C. Koch, and E. Niebur, "A model of saliency-based visual attention for rapid scene analysis," *IEEE Trans. Pattern Anal. Mach. Intell.,* vol. 20, no. 11, pp. 1254-1259, 1998.

[3] M. Liang, and X. Hu, "Predicting eye fixations with higher-level visual features," *IEEE Trans. Image Process.,* vol. 24, no. 3, pp. 1178-1189, 2015.

[4] T. Judd, K. Ehinger, F. Durand, and A. Torralba, "Learning to predict where humans look," in *Proc. IEEE Int. Conf. Comput. Vision*, 2009, pp. 2106-2113.

[5] A. Borji, "Boosting bottom-up and top-down visual features for saliency estimation," in *Proc. IEEE Conf. Comput. Vision Pattern Recogn.*, 2012, pp. 438-445.

[6] C. Guo, and L. Zhang, "A novel multiresolution spatiotemporal saliency detection model and its applications in image and video compression," *IEEE Trans. Image Process.,* vol. 19, no. 1, pp. 185-198, 2010.

[7] M. W. Cannon, and S. C. Fullenkamp, "A model for inhibitory lateral interaction effects in perceived contrast," *Vis. Res.,* vol. 36, no. 8, pp. 1115-1125, 1996.

[8] F. Crick, and C. Koch, "Towards a neurobiological theory of consciousness," *Semin. Neurosci.,* vol. 2, pp. 263-275, 1990.

[9] E. Niebur, C. Koch, and C. Rosin, "An oscillation-based model for the neuronal basis of attention," *Vis. Res.,* vol. 33, no. 18, pp. 2789-2802, 1993.

[10] J. Harel, C. Koch, and P. Perona, "Graph-based visual saliency," in *Proc. Adv. Neural Inform. Process. Syst.*, 2006, pp. 545-552.

[11] Y. LeCun, L. Bottou, Y. Bengio, and P. Haffner, "Gradient-based learning applied to document recognition," *Proc. IEEE,* vol. 86, no. 11, pp. 2278-2324, 1998.

[12] K. Simonyan, and A. Zisserman, "Very Deep Convolutional Networks for Large-Scale Image

Recognition," in *Proc. Int. Conf. Learn. Representations*, 2015.

[13] K. He, X. Zhang, S. Ren, and J. Sun, "Deep residual learning for image recognition," *arXiv preprint arXiv:1512.03385*, 2015.

[14] J. Deng, W. Dong, R. Socher, L. J. Li, K. Li, and F. F. Li, "ImageNet: A large-scale hierarchical image database," in *Proc. IEEE Conf. Comput. Vision Pattern Recogn.*, 2009, pp. 248-255.

[15] S. Hochreiter, and J. Schmidhuber, "Long short-term memory," *Neural Comput.*, vol. 9, no. 8, pp. 1735-1780, 1997.

[16] A. Torralba, A. Oliva, M. S. Castelhano, and J. M. Henderson, "Contextual guidance of eye movements and attention in real-world scenes: the role of global features in object search," *Psychol. Rev.*, vol. 113, no. 4, pp. 766, 2006.

[17] B. C. Russell, A. Torralba, K. P. Murphy, and W. T. Freeman, "LabelMe: a database and web-based tool for image annotation," *Int. J. Comput. Vision*, vol. 77, no. 1-3, pp. 157-173, 2008.

[18] B. Zhou, A. Lapedriza, J. Xiao, A. Torralba, and A. Oliva, "Learning deep features for scene recognition using places database," in *Proc. Adv. Neural Inform. Process. Syst.*, 2014, pp. 487-495.

[19] S. Ghosh, O. Vinyals, B. Strope, S. Roy, T. Dean, and L. Heck, "Contextual LSTM (CLSTM) models for Large scale NLP tasks," *arXiv preprint arXiv:1602.06291*, 2016.

[20] N. Bruce, and J. Tsotsos, "Saliency based on information maximization," in *Proc. Adv. Neural Inform. Process. Syst.*, 2005, pp. 155-162.

[21] D. Gao, V. Mahadevan, and N. Vasconcelos, "On the plausibility of the discriminant center-surround hypothesis for visual saliency," *J. Vis.*, vol. 8, no. 7, pp. 13, 2008.

[22] H. J. Seo, and P. Milanfar, "Static and space-time visual saliency detection by self-resemblance," *J. Vis.*, vol. 9, no. 12, pp. 15, 2009.

[23] B. Han, H. Zhu, and Y. Ding, "Bottom-up saliency based on weighted sparse coding residual," in *Proc. ACM Int. Conf. Multimedia*, 2011, pp. 1117-1120.

[24] X. Hou, and L. Zhang, "Saliency detection: A spectral residual approach," in *Proc. IEEE Conf. Comput. Vision Pattern Recogn.*, 2007, pp. 1-8.

[25] L. Zhang, M. H. Tong, T. K. Marks, H. Shan, and G. W. Cottrell, "SUN: A Bayesian framework for saliency using natural statistics," *J. Vis.*, vol. 8, no. 7, pp. 32, 2008.

[26] X. Hou, J. Harel, and C. Koch, "Image signature: Highlighting sparse salient regions," *IEEE Trans. Pattern Anal. Mach. Intell.*, vol. 34, no. 1, pp. 194-201, 2012.

[27] A. Garcia-Diaz, X. R. Fdez-Vidal, X. M. Pardo, and R. Dosil, "Saliency from hierarchical adaptation through decorrelation and variance normalization," *Image Vis. Comput.*, vol. 30, no. 1, pp. 51-64, 2012.

[28] A. Krizhevsky, I. Sutskever, and G. E. Hinton, "Imagenet classification with deep convolutional neural networks," in *Proc. Adv. Neural Inform. Process. Syst.*, 2012, pp. 1097-1105.

[29] R. Girshick, J. Donahue, T. Darrell, and J. Malik, "Rich feature hierarchies for accurate object detection and semantic segmentation," in *Proc. IEEE Conf. Comput. Vision Pattern Recogn.*, 2014, pp. 580-587.

[30] J. Long, E. Shelhamer, and T. Darrell, "Fully Convolutional Networks for Semantic Segmentation," in *Proc. IEEE Conf. Comput. Vision Pattern Recogn.*, 2015, pp. 3431-3440.

[31] G. Li, and Y. Yu, "Visual Saliency Detection Based on Multiscale Deep CNN Features," *IEEE Trans. Image Process.*, vol. 25, no. 11, pp. 5012-5024, 2016.

[32] L. Wang, H. Lu, X. Ruan, and M.-H. Yang, "Deep Networks for Saliency Detection via Local Estimation and Global Search," in *Proc. IEEE Conf. Comput. Vision Pattern Recogn.*, 2015, pp. 3183-3192.

[33] R. Zhao, W. Ouyang, H. Li, and X. Wang, "Saliency detection by multi-context deep learning," in *Proc. IEEE Conf. Comput. Vision Pattern Recogn.*, 2015, pp. 1265-1274.

[34] N. Liu, and J. Han, "DHSNet: Deep Hierarchical Saliency Network for Salient Object Detection," in *Proc. IEEE Conf. Comput. Vision Pattern Recogn.*, 2016, pp. 678-686.

[35] C. Shen, M. Song, and Q. Zhao, "Learning high-level concepts by training a deep network on eye fixations," in *Proc. NIPS Deep Learn. Unsupervised Feature Learn. Workshop*, 2012.

[36] E. Vig, M. Dorr, and D. Cox, "Large-Scale Optimization of Hierarchical Features for Saliency Prediction in Natural Images," in *Proc. IEEE Conf. Comput. Vision Pattern Recogn.*, 2014, pp. 2798–2805.

[37] J. Han, D. Zhang, S. Wen, L. Guo, T. Liu, and X. Li, "Two-stage learning to predict human eye fixations via SDAEs," *IEEE Trans. Cybern.*, vol. 46, no. 2, pp. 487-498, 2016.

[38] M. Kümmerer, L. Theis, and M. Bethge, "Deep Gaze I: Boosting Saliency Prediction with Feature Maps Trained on ImageNet," *arXiv preprint arXiv:1411.1045*, 2014.

[39] N. Liu, J. Han, D. Zhang, S. Wen, and T. Liu, "Predicting Eye Fixations using Convolutional Neural Networks," in *Proc. IEEE Conf. Comput. Vision Pattern Recogn.*, 2015, pp. 362-370.

[40] X. Huang, C. Shen, X. Boix, and Q. Zhao, "SALICON: Reducing the Semantic Gap in Saliency Prediction by Adapting Deep Neural Networks," in *Proc. IEEE Int. Conf. Comput. Vision*, 2015, pp. 262-270.

[41] S. S. S. Kruthiventi, K. Ayush, and R. V. Babu, "DeepFix: A Fully Convolutional Neural Network for predicting Human Eye Fixations," *arXiv preprint arXiv:1510.02927*, 2015.

[42] S. S. S. Kruthiventi, V. Gudisa, J. H. Dholakiya, and R. V. Babu, "Saliency Unified: A Deep Architecture for Simultaneous Eye Fixation Prediction and Salient Object Segmentation," in *Proc. IEEE Conf. Comput. Vision Pattern Recogn.*, 2016, pp. 5781-5790.

[43] J. Pan, K. McGuinness, E. Sayrol, N. O'Connor, and X. Giro-i-Nieto, "Shallow and Deep Convolutional Networks for Saliency Prediction," in *Proc. IEEE Conf. Comput. Vision Pattern Recogn.*, 2016, pp. 598-606.

[44] S. Jetley, N. Murray, and E. Vig, "End-to-End Saliency Mapping via Probability Distribution Prediction," in *Proc. IEEE Conf. Comput. Vision Pattern Recogn.*, 2016, pp. 5753-5761.

[45] M. Cornia, L. Baraldi, G. Serra, and R. Cucchiara, "A Deep Multi-Level Network for Saliency Prediction," in *Proc. Int. Conf. Pattern Recogn.*, 2016.

[46] A. Graves, and N. Jaitly, "Towards End-To-End Speech Recognition with Recurrent Neural Networks," in *Proc. Int. Conf. Mach. Learn.*, 2014, pp. 1764-1772.

[47] I. Sutskever, O. Vinyals, and Q. V. Le, "Sequence to sequence learning with neural networks," in *Proc. Adv. Neural Inform. Process. Syst.*, 2014, pp. 3104-3112.

[48] J. Donahue, L. Anne Hendricks, S. Guadarrama, M. Rohrbach, S. Venugopalan, K. Saenko, and T. Darrell, "Long-term recurrent convolutional networks for visual recognition and description," in *Proc. IEEE Conf. Comput. Vision Pattern Recogn.*, 2015, pp. 2625-2634.

[49] F. Visin, K. Kastner, K. Cho, M. Matteucci, A. Courville, and Y. Bengio, "Renet: A recurrent neural network based alternative to convolutional networks," *arXiv preprint arXiv:1505.00393*, 2015.

[50] S. Bell, C. L. Zitnick, K. Bala, and R. Girshick, "Inside-outside net: Detecting objects in context with skip pooling and recurrent neural networks," *arXiv preprint arXiv:1512.04143*, 2015.

[51] Z. Yan, H. Zhang, Y. Jia, T. Breuel, and Y. Yu, "Combining the Best of Convolutional Layers and Recurrent Layers: A Hybrid Network for Semantic Segmentation," *arXiv preprint arXiv:1603.04871*, 2016.

[52] F. Yu, and V. Koltun, "Multi-scale context aggregation by dilated convolutions," *arXiv preprint arXiv:1511.07122*, 2015.

[53] V. Nair, and G. E. Hinton, "Rectified linear units improve restricted boltzmann machines," in *Proc. Int. Conf. Mach. Learn.*, 2010, pp. 807-814.

[54] W. Liu, A. Rabinovich, and A. C. Berg, "Parsenet: Looking wider to see better," *arXiv preprint arXiv:1506.04579*, 2015.

[55] S. Ioffe, and C. Szegedy, "Batch normalization: Accelerating deep network training by reducing internal covariate shift," *arXiv preprint arXiv:1502.03167*, 2015.

[56] P. J. Werbos, "Backpropagation through time: what it does and how to do it," *Proc. IEEE,* vol. 78, no. 10, pp. 1550-1560, 1990.

[57] A. Graves, N. Jaitly, and A.-r. Mohamed, "Hybrid speech recognition with deep bidirectional LSTM," in *Proc. ASRU Workshop*, 2013, pp. 273-278.

[58] R. J. Peters, A. Iyer, L. Itti, and C. Koch, "Components of bottom-up gaze allocation in natural images," *Vis. Res.,* vol. 45, no. 18, pp. 2397-2416, 2005.

[59] Z. Bylinskii, T. Judd, A. Oliva, A. Torralba, and F. Durand, "What do different evaluation metrics tell us about saliency models?," *arXiv preprint arXiv:1604.03605*, 2016.

[60] D. E. Rumelhart, G. E. Hinton, and R. J. Williams, "Learning internal representations by error propagation," *Parallel Distributed Processing,* vol. 1, pp. 318–362, 1986.

[61] M. Jiang, S. Huang, J. Duan, and Q. Zhao, "SALICON: Saliency in context," in *Proc. IEEE Conf. Comput. Vision Pattern Recogn.*, 2015, pp. 1072-1080.

[62] T.-Y. Lin, M. Maire, S. Belongie, J. Hays, P. Perona, D. Ramanan, P. Dollár, and C. L. Zitnick, "Microsoft coco: Common objects in context," in *Proc. Eur. Conf. Comput. Vision*, 2014, pp. 740-755.

[63] T. Judd, F. Durand, and A. Torralba, "A benchmark of computational models of saliency to predict human fixations," *MIT Technical Report*, 2012.

[64] B. W. Tatler, R. J. Baddeley, and I. D. Gilchrist, "Visual correlates of fixation selection: effects of scale and time," *Vis. Res.,* vol. 45, no. 5, pp. 643-659, 2005.

[65] Y. Jia, E. Shelhamer, J. Donahue, S. Karayev, J. Long, R. Girshick, S. Guadarrama, and T. Darrell, "Caffe: Convolutional architecture for fast feature embedding," in *Proc. ACM Int. Conf. Multimedia*, 2014, pp. 675-678.

[66] J. Zhang, and S. Sclaroff, "Saliency detection: a boolean map approach," in *Proc. IEEE Int. Conf. Comput. Vision*, 2013, pp. 153-160.

[67] Q. Zhao, and C. Koch, "Learning a saliency map using fixated locations in natural scenes," *J. Vis.,* vol. 11, no. 3, pp. 9, 2011.

[68] A. Borji, D. N. Sihite, and L. Itti, "Quantitative analysis of human-model agreement in visual saliency modeling: a comparative study," *IEEE Trans. Image Process.,* vol. 22, no. 1, pp. 55-69, 2013.